\documentclass[conference, 10 pt, conference]{IEEEtran}
\usepackage[T1]{fontenc}
\usepackage{aecompl} 
\usepackage{caption}
\IEEEoverridecommandlockouts
\usepackage{cite}
\usepackage{amsmath,amssymb,amsfonts}
\usepackage{algorithm}
\usepackage{algorithmic}
\usepackage{graphicx}
\usepackage{textcomp}
\usepackage{xcolor}
\usepackage{diagbox}
\usepackage{tabu}
\usepackage{subfigure}
\usepackage{multirow}

\usepackage{algorithmic}
\usepackage[colorlinks,
            linkcolor=black,       
            anchorcolor=black,  
            citecolor=black,
            urlcolor=black,        
            ]{hyperref}

\usepackage[letterpaper, 
            left=19.1mm, 
            right=19.1mm, 
            top=19.1mm, 
            bottom=19.1mm]{geometry}

\def\BibTeX{{\rm B\kern-.05em{\sc i\kern-.025em b}\kern-.08em
    T\kern-.1667em\lower.7ex\hbox{E}\kern-.125emX}}
\begin{document}

\title{\Huge SmartCooper: Vehicular Collaborative Perception with Adaptive Fusion and Judger Mechanism}

\author{
    \IEEEauthorblockN{Yuang Zhang${}^{1}$, Haonan An${}^{2}$, Zhengru Fang${}^{3}$, Guowen Xu${}^{3}$, Yuan Zhou${}^{2}$, Xianhao Chen${}^{4}$, Yuguang Fang${}^{3}$}
}

\maketitle

\newcommand\blfootnote[1]{%
\begingroup
\renewcommand\thefootnote{}\footnote{#1}%
\addtocounter{footnote}{-1}%
\endgroup
}

\blfootnote{${}^1$Department of Automation, Tsinghua University. \texttt{zya21@mails.tsi nghua.edu.cn}}
\blfootnote{${}^2$School of Electrical and Electronic Engineering, Nanyang Technological University. \texttt{\{an0029an, yzhou027\}@e.ntu.edu.sg}}
\blfootnote{${}^3$Department of Computer Science, City University of Hong Kong. \texttt{zhef ang4-c@my.cityu.edu.hk, \{guowenxu, my.Fang\}@cityu.e du.hk}}
\blfootnote{${}^4$Department of Electrical and Electronic Engineering, The University of Hong Kong. \texttt{xchen@eee.hku.hk}}

\vspace{-0.35cm}

\begin{abstract}
In recent years, autonomous driving has garnered significant attention due to its potential for improving road safety through collaborative perception among connected and autonomous vehicles (CAVs). However, time-varying channel variations in vehicular transmission environments demand dynamic allocation of communication resources. Moreover, in the context of collaborative perception, it is important to recognize that not all CAVs contribute valuable data, and some CAV data even have detrimental effects on collaborative perception. In this paper, we introduce SmartCooper, an adaptive collaborative perception framework that incorporates communication optimization and a judger mechanism to facilitate CAV data fusion. Our approach begins with optimizing the connectivity of vehicles while considering communication constraints. We then train a learnable encoder to dynamically adjust the compression ratio based on the channel state information (CSI). Subsequently, we devise a judger mechanism to filter the detrimental image data reconstructed by adaptive decoders. We evaluate the effectiveness of our proposed algorithm on the OpenCOOD platform. Our results demonstrate a substantial reduction in communication costs by 23.10\% compared to the non-judger scheme. Additionally, we achieve a significant improvement on the average precision of Intersection over Union (AP@IoU) by 7.15\% compared with state-of-the-art schemes.
\end{abstract}


\section{Introduction}
Over the past few decades, autonomous driving perception technology has witnessed remarkable progress, thanks to the rapid advances of computer vision and deep learning techniques\cite{chen2023vaas}\cite{hu2024collaborative}. This technology relies predominantly on two types of data sources: image data and point cloud data, both acquired from on-board sensors in real-world scenarios or generated through simulation platforms\cite{zhang2023quality}. In terms of real-world data, researchers have proposed extensive public datasets like KITTI\cite{geiger2013kitti} and nuScenes\cite{caesar2020nuscenes}. By using simulation platforms like CARLA\cite{dosoviskiy2017carla}, simulation datasets like OPV2V\cite{xu2022opv2v} and SHIFT\cite{sun2022shift}, usually include a wider range of scenarios and weather conditions, along with comprehensive annotation information, thereby facilitating the studies of autonomous driving perception. Numerous single-vehicle perception methods have been rigorously tested and validated on both real-world and simulation datasets\cite{chen2023vaas},\cite{lang2019pointpillars},\cite{xu2018pointfusion}. However, recent developments in communication technology and the advent of CAVs have thrusted collaborative perception into the forefront of autonomous driving research\cite{talebpour2016review}\cite{hu2023towards}.  In contrast to single-vehicle perception, collaborative perception extends the scope of perception and effectively mitigates blind spots\cite{wei2022threeu}. This allows autonomous vehicles to comprehensively perceive their surroundings, leading to more effective decision-making\cite{cui2022review}. Specifically, collaborative perception systems facilitate the sharing of perception data among CAVs\cite{linz2022tracking}. Each CAV benefits from a broader sensing range than single-vehicle perception, thus aggregating global and local information to optimize perception results\cite{kim2014daev}\cite{hu2023adaptive}. Existing studies have introduced various collaborative perception methods, including a perception framework based on point cloud features\cite{chen2019fcooper} and a collaborative perception framework based on Bird's-Eye View (BEV), which refers to the top view of a scene, and vision\cite{xu2023cobevt}.

Nonetheless, several challenges persist in current collaborative perception research, particularly when applied to real-world scenarios. Firstly, the data collected by vehicular sensors, comprising image sequences and point cloud data, often pose a significant volume challenge. Regarding image data, contemporary cameras, exemplified by Google's autonomous vehicles (AVs), are capable of capturing a staggering 750 megabytes of data per second\cite{hu2019dnn}. As for point cloud data, using KITTI\cite{geiger2013kitti} as an illustration, even the smallest sequence contains upwards of 10 million data points\cite{liy2020lidarreview}. The sheer volume of data can easily congest communication channels and potentially result in significant delays or packet loss, thereby restricting effective data sharing among CAVs. Moreover, the real-world communication landscape is far from ideal, and with constrained spectrum resources is difficult to enable real-time data transmissions in Vehicle-to-Vehicle (V2V) networks\cite{chen2023vaas}. Specifically, channel quality fluctuates over time, and the limited bandwidth cannot guarantee consistent real-time and complete data transmission from all vehicles. Some prior studies have already recognized these challenges and initiated efforts to address them. For instance, Hu et al. proposed an efficient communication framework, Where2comm, incorporating an attention-based mechanism to enhance perception performance and mitigate bandwidth pressures\cite{hu2022where2comm}. Similarly, Wang et al. introduced a viable approach wherein each agent transmits compressed feature maps, successfully alleviating communication burdens\cite{wang2020v2vnet}. However, these approaches often assume that channel condition remains time-invariant and fail to account for dynamic channel variations. In contrast, channel-aware methods prioritize the real-time fluctuations in channel conditions among CAVs, adjusting data transmissions accordingly. This dynamic adaptation optimizes communication, ensuring the efficacy of collaborative perception.

Secondly, data compression for CAVs also affects the performance of collaborative perception significantly. However, existing data compression schemes exhibit notable deficiencies. Previous compression methods have typically employed a uniform ``fairness scheme'' \cite{magsino2022roadside}, which uniformly applies the same compression ratio to all CAV data. Unfortunately, this fairness scheme overlooks the fact that CAVs in a V2V network contribute perception data differently in data fusion. Some CAVs, situated at some locations, can significantly extend sensing coverage and enhance collaborative perception, while others, with obscured views, provide minimal assistance to data fusion. Moreover, it is crucial to recognize that not all data from CAVs is beneficial. Certain CAVs may be situated at the edge of perception or suffer from severe communication delays, potentially having a detrimental impact on data fusion and consequently reducing the accuracy of collaborative perception. Thus, a judger mechanism is needed to assess and prioritize CAVs based on their spatial location and sensing coverage. This mechanism can then eliminate low-quality data, enabling efficient CAV data fusion and perception.

Finally, most existing collaborative perception schemes assume that the wireless channel is ideal, ignoring the dynamic characteristics of the channel and the mechanism of filtering the perception data. If the dynamic nature of the channel is taken into account, the compression ratio can be flexibly adjusted and the communication can be optimized. Moreover, the priority can be set adaptively according to the sensing coverage and data quality of CAVs. In contrast to the previous studies, in this paper, we propose SmartCooper, an adaptive collaborative perception framework with channel-aware communication optimization and judger mechanism to enable efficient CAVs’ data fusion. Our main contributions are summarized as follows. 

\begin{itemize}
\item To the best of our knowledge, SmartCooper is the first that comprehensively considers CAV data judger mechanism. We design a novel judger to filter reconstructed image data according to the sensing coverage. Specifically, our judger scores reconstructed image sequences based on the union of sensing coverage, then removes data with negative gains to improve the accuracy of collaborative perception and reduce communication cost.
\item We propose a channel-aware communication optimizing scheme to adapt to the channel state. Specifically, we leverage an adaptive encoder to adjust the compression ratio of data delivered by CAVs according to the real-time channel state, ensuring real-time and high-quality data transmission for CAV data fusion. 
\item With the OpenCOOD platform\cite{xu2022opv2v}, we conduct extensive experiments to validate the effectiveness of our proposed scheme by comparing it with the state-of-the-art methods. Under the same channel condition, the communication cost shows a significant reduction by at least 23.10\%, and AP@IoU (the average precision of Intersection over Union), which evaluates the accuracy of perception, is significantly improved by 7.15\%.
\end{itemize}

\section{Related Work}
\subsection{Collaborative Perception}
Due to the limited sensing range of a vehicle's sensors, CAVs are unable to sense the imminent dangers behind certain hard-to-observe corners. However, collaboration with other CAVs can expand its sensing coverage, thus reducing safety risks. Numerous works on collaborative perception have already been carried out lately. Wang \textit{et al.}\cite{wang2020v2vnet} proposed V2V collaboration based on point cloud data, while Xu \textit{et al.}\cite{xu2023cobevt} introduced V2V collaboration using 2D image data. Although these studies have demonstrated promising results on test datasets, they often assume an ideal channel environment for V2V collaboration. When these methods are applied in real-world scenarios, the presence of non-ideal communication channels will cause serious problems such as long latency and high packet loss because of the resulting reduced transmission rate. Thus, while these approaches have shown potential, their effectiveness under realistic channel conditions needs further consideration.
\subsection{Which Vehicle to Communicate}
The choice of vehicles to participate in collaboration not only significantly impacts spectrum consumption but also the quality of fused sensing results, thereby affecting the overall performance of cooperation. Therefore, the selection of which vehicles to participate is of great importance. Liu \textit{et al.}\cite{liu2020who2com} proposed a three-stage handshake mechanism for communication selection. Specifically, participating vehicles are chosen based on matching scores computed between the data from the degraded agent and data from all other normal agents. Hu \textit{et al.}\cite{hu2022where2comm} suggested establishing a spatial confidence map for participating vehicle selection. However, most of these methods heavily rely on attention mechanisms to build an evaluation system. Assuming a data volume with $N$ participating vehicles, attention mechanisms implies a time complexity of $O(N^2)$  for communications, a non-negligible cost that could result in additional latency.

\section{Our Proposed Scheme}
\label{sec:proposed_method}
In this section, we introduce our proposed adaptive collaborative scheme for CAVs. Firstly, in Sec. \ref{sec:channel-aware}, we mathematically model the channel conditions in practical communication scenarios, thus defining the adaptive factors required for CAVs in practical settings. Next, in Sec. \ref{sec:weakness}, we conduct a concise experiment to illustrate the drawbacks of solely optimizing communication parameters for enhancing collaborative quality. Then, in Sec. \ref{sec:priority}, we introduce a scoring mechanism based on the perceptual gains of the Ego CAV to address the shortcomings mentioned in Sec. \ref{sec:weakness}, that is, some CAVs may have detrimental effects on sensing data fusion. Finally, in Sec. \ref{sec:overall architecture} we combine the two adaptive strategies and show our comprehensive network architecture.

\subsection{Channel-aware Optimization}
\label{sec:channel-aware}
\subsubsection{CAV Communication Scheme}
In view of the fact that the actual communication situation is not ideal, we propose a channel-aware communication optimizing scheme. We formulate the channel capacity optimization under CAVs' actual communication conditions, and propose our adaptive communication scheme. In accordance with the 5G standards by 3GPP\cite{anwar2019threegpp},\cite{fang2022ageofinfo}, CAVs primarily rely on Cellular Vehicle-to-Everything (C-V2X) employing Orthogonal Frequency Division Multiplexing (OFDM). The capacity of each sub-channel ${SC}_{ij}$ is given by:

\begin{equation}
{SC}_{ij} = \frac{W}{N} \log_2 { \left[ 1 + \frac{H_{ij} * P_{T}}{\rho_{n} * \left( \frac{W}{N} \right)} \right] },
\label{eq:communicaiton volumn}
\end{equation} 
where $W$ denotes the total bandwidth, $N$ the number of sub-channels, ${SC}_{ij}$ the sub-channel capacity from the $i$th transmitter to the $j$th receiver, $H_{ij}$ the channel gain from the $i$th transmitter to the $j$th receiver, $P_{T}$ the transmit power, and $\rho_{n}$ the noise power spectral density.

\subsubsection{Channel-aware Adjuster}
Based on the above modeling, we get the sub-channel capacity $SC_{ij}$. Moreover, we propose a channel-aware adjuster to adaptively adjust the compression ratio of each agent's data according to the communication capacity of the sub-channel. The compression ratio $\alpha^{*}_{ij}$ of each sub-channel is given by:

\begin{equation}
\alpha^{*}_{ij} = \frac{{SC}_{ij}}{V_{ij}}, \label{eq: compression ratio}
\end{equation}
where ${\alpha^*}_{ij}$ is the compression ratio of the data transmitted from the $i$th transmitter to the $j$th receiver, and $V_{ij}$ is the volume of data transmitted from the $i$th transmitter to the $j$th receiver. Since the volume of data transmitted is often greater than the capacity of a single sub-channel, the compression ratio satisfies: $0 \leq {\alpha^*}_{ij} \leq 1, {\alpha^*}_{ij} \in R ^ { + }$. Based on this, the adaptive reconstruction structure is designed as follows: we utilize channel-aware adjuster to modulate the compression ratio of the transmitted data for each CAV according to the capacity of the sub-channel, then the encoder transmits the compressed data, and the decoder receives the information transmitted over the sub-channel and enables the reconstruction.

\subsection{Adaptive Judger Mechanism}

\begin{table}
\centering
\renewcommand{\arraystretch}{1.2}
\caption{The collaborative results for different numbers of CAVs under both ideal and non-ideal conditions.}
\label{table:weakness}
\resizebox{0.49\textwidth}{!}{
\begin{tabular}{|c|cccc|}
\hline
\multirow{2}{*}{\diagbox[width=10em]{AP@IoU}{Parameters}} & \multicolumn{4}{c|}{\#CAVs}                                                  \\ \cline{2-5} 
                  & \multicolumn{1}{c|}{1} & \multicolumn{1}{c|}{2} & \multicolumn{1}{c|}{3} & \multicolumn{1}{c|}{4} \\ \hline
Ideal             & \multicolumn{1}{c|}{0.252}  & \multicolumn{1}{c|}{0.582}  & \multicolumn{1}{c|}{0.668} & \multicolumn{1}{c|}{0.673}    \\ \hline
Non-ideal         & \multicolumn{1}{c|}{0.252}  & \multicolumn{1}{c|}{0.579}  & \multicolumn{1}{c|}{0.656} & \multicolumn{1}{c|}{0.642}   \\ \hline
\end{tabular}
}
\vspace{-0.2cm}
\end{table}

\subsubsection{Disadvantage Analysis}
\label{sec:weakness}
The proposed method in Sec. \ref{sec:channel-aware} can adaptively allocate communication resources to different CAVs based on the channel conditions, thereby enhancing the performance of V2V collaborative perception. However, we have observed that the optimized collaborative data may not always lead to positive gains (i.e., it might potentially degrade the perception performance of the Ego CAV). To illustrate this issue, we conducted a straightforward yet convincing experiment. The results are summarized in Table \ref{table:weakness}.

In our experiment, we employ a test dataset from OPV2V\cite{xu2022opv2v} involving a collaborative network of four CAVs. To systematically analyze the impact of participating CAVs on collaborative perception, we increase the number of collaborating CAVs incrementally based on their distance to the Ego CAV. Notably, when the number of CAVs equals 1, it signifies the presence of only the Ego CAV, with no collaborative perception taking place. We conducted tests to measure AP@IoU for the collaborative network under two distinct scenarios: ideal condition and non-ideal condition (Transmissions under non-ideal condition are achieved by adaptively allocating channel resources using the approach outlined in Sec. \ref{sec:channel-aware}). Under ideal channel conditions, data transmission does not require any compression. When increasing the number of CAVs, AP@IoU also increases simultaneously. This occurs because different CAVs expand the Ego CAV's sensing coverage, enhancing its understanding of the surrounding environment. However, under non-ideal channel conditions, an interesting phenomenon emerges when increasing the number of CAVs. Initially, AP@IoU increases, but then it starts to decrease. It indicates that not all CAVs contribute positively under non-ideal conditions. Therefore, we conclude that optimizing the channel under non-ideal conditions alone is insufficient to achieve optimal collaborative performance.

\begin{figure}[t]
  \centering
  \subfigure[Score Increasing]{
  \includegraphics[width=4.0cm]{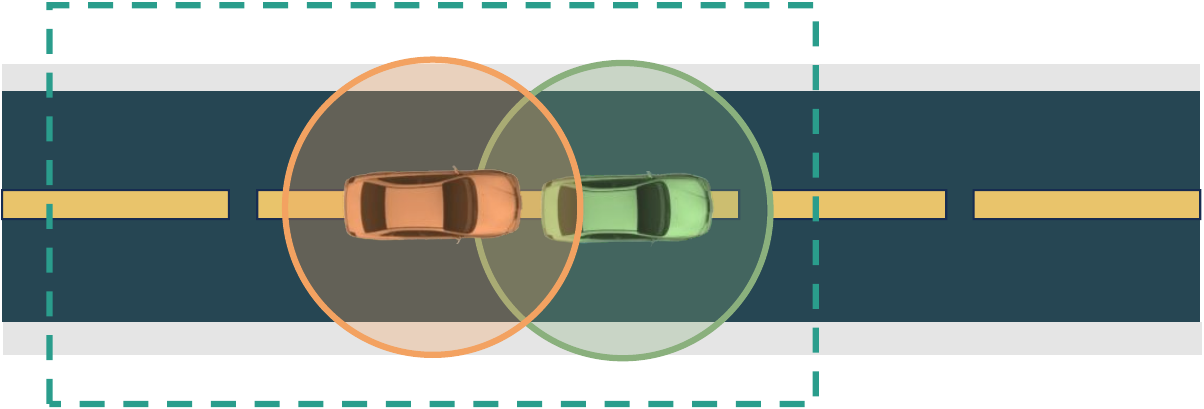}\label{fig:EE-TC-1}
  }
  \subfigure[Maximum Score]{
  \includegraphics[width=4.0cm]{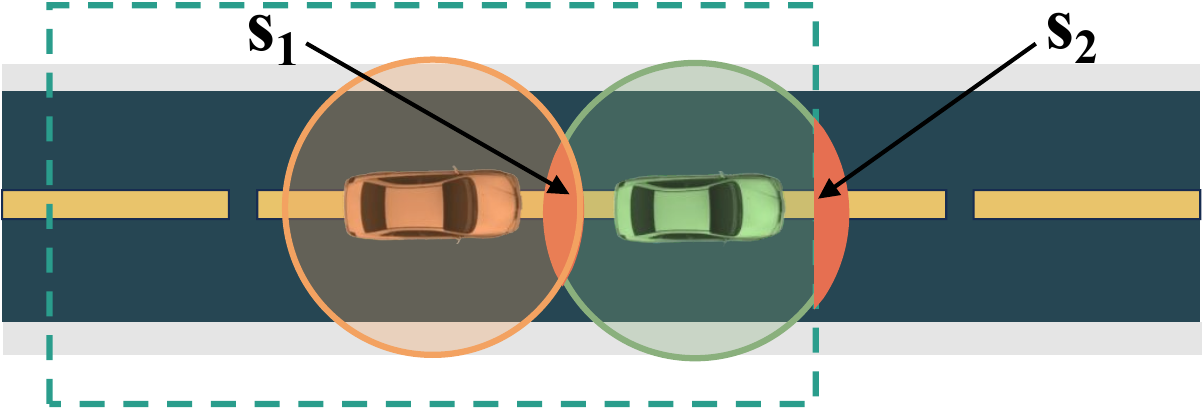}\label{fig:EE-TC-2}
  }
  \subfigure[Score Decreasing]{
  \includegraphics[width=4.0cm]{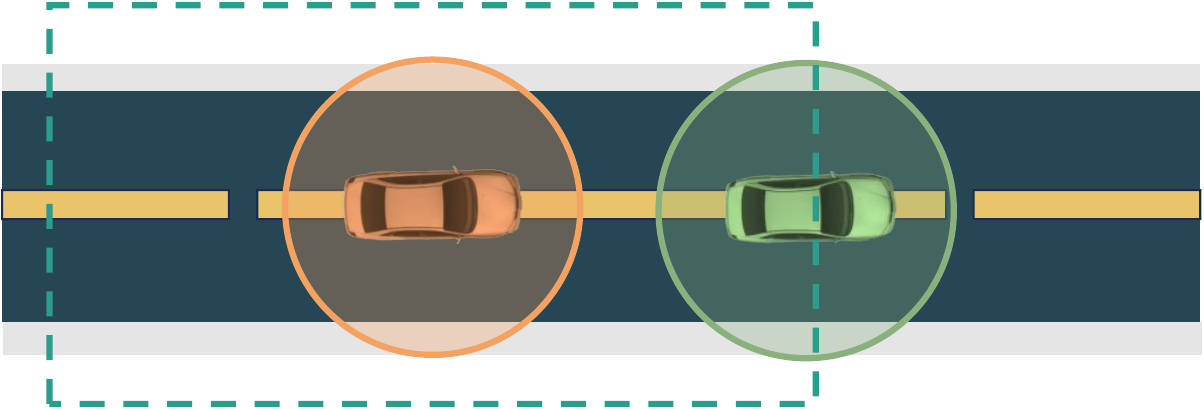}\label{fig:EE-TC-3}
  }
  \subfigure[Score equal to zero]{
  \includegraphics[width=4.0cm]{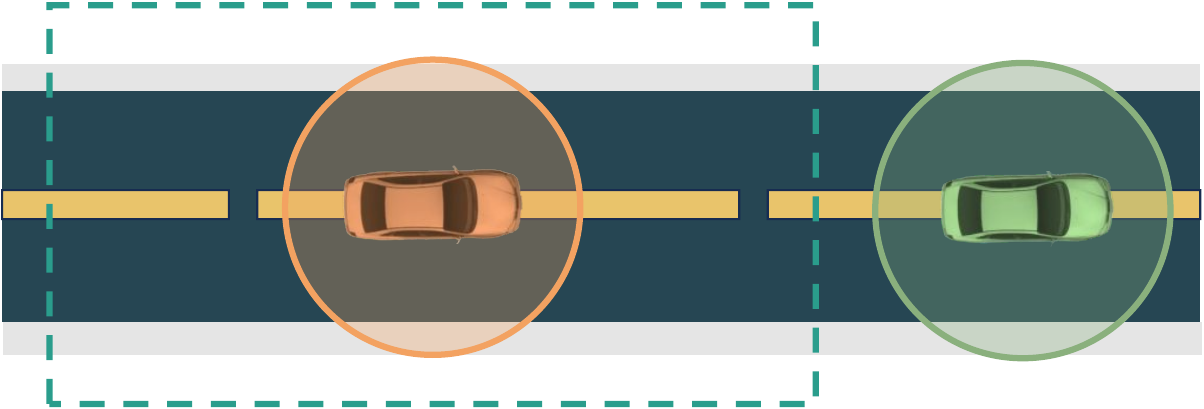}\label{fig:EE-TC-4}
  }
  \caption{Four possible scoring scenarios where the orange represents the Ego CAV, and the green represents CAV1.}
  \label{fig:sening_range}
  \vspace{-0.3cm}
\end{figure}

\begin{figure*}[t]
  \centering
  \includegraphics[width=2\columnwidth]{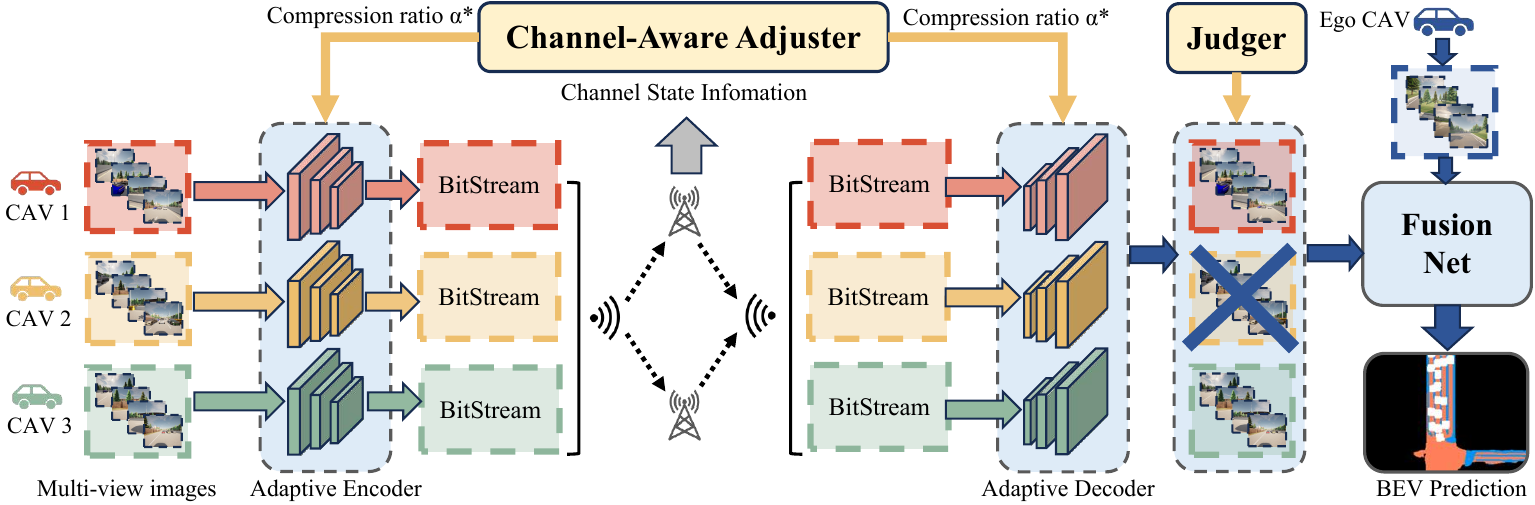}
  \caption{Overall architecture of SmartCooper consists of three stages. (1) Network optimization: determining the parameters of CSI to output the optimized compression ratio $\alpha^{*}$. (2) Transmission \& Judgement: Each CAV transmits its compressed image data to the Ego CAV and receives a judgment. (3) Fusion: the qualified reconstructed images are combined with the Ego CAV's image to generate a BEV through the fusion net.}
  \label{fig:overall_architecture}
  \vspace{-0.3cm}
\end{figure*}

\subsubsection{Scoring \& Threshold Scheme}
\label{sec:priority}
In this subsection, we introduce our judger mechanism based on scoring and threshold to mitigate the detrimental impact of the data with negative gain on collaborative quality. Specifically, as for the Ego CAV, we define a predetermined maximum collaborative sensing coverage (MCSC) as $d_{max}$. Additionally, for each CAV, we define the individual sensing coverage denoted as $d_{single}$ for each CAV, including the ego CAV, as follows:
\begin{equation}
    d_{single} < d_{max}.
\end{equation}
Hence, the sensing coverage of each individual CAV can be obtained as follows:
\begin{equation}
    s = \pi \times d^{2}_{single} - s_{otr},
\end{equation}
where $s_{otr}$ represents the area out of MCSC. Meanwhile, we define the score of each CAV, or rather, the collaborative gain of the Ego CAV, in the following equation:
\begin{equation}
    \label{eq:score}
    score = s_{cav}\bigcup s_{ego} - s_{ego}.
\end{equation}
where $s_{cav}$ denotes the sensing coverage of each CAV, and $s_{ego}$ denotes the sensing coverage of ego CAV. Intuitively, the score captures the sensing area of a collaborative CAV that the ego CAV cannot sense. The different situations in which the collaborative gain varies with sensing coverage among CAVs are shown in Fig. \ref{fig:sening_range}. For the sake of analysis, we assume that the Ego CAV remains stationary. As CAV moves, the variations in the collaborative gain of the Ego CAV exhibit four scenarios: (a) When the distance between the CAV and the Ego CAV is short, the collaborative gain of the Ego CAV increases as the CAV approaches. This trend continues until (b) The intersection of the CAV's sensing coverage with the Ego CAV's sensing coverage equals the intersection of the CAV's sensing coverage with MCSC. In other words, $s_{1} = s_{2}$. (c) As the CAV continues to move, the collaborative gain of the Ego CAV gradually decreases due to most of perception part exceeding MCSC until (d) The CAV moves entirely out of MCSC.

According to Eq. (\ref{eq:communicaiton volumn}) and Eq. (\ref{eq: compression ratio}), the farther a CAV is from the Ego CAV within the MCSC, the fewer channel resources it will be allocated. Additionally, CAVs located far from the Ego CAV obtain low scores based on Eq. (\ref{eq:score}). These two factors are combined to find CAVs near the vicinity of the MCSC contributing poorer quality and less impactful data to the collaboration with the Ego CAV. Consequently, lack of the consideration of the impact of the detrimental data can lead to interference or even errors. This phenomenon explains the situation observed in cases when an increased number of participating CAVs results in a reduction in collaborative performance in Table \ref{table:weakness}. Hence, guided by experimental data, we have established a threshold based on the experience. This threshold serves to mitigate the adverse impact of detrimental data on collaborative quality. Data falling below the established threshold is discarded outright, while data surpassing the threshold is retained for further processing. It is worth noting that, compared to the previous selection strategies based on attention mechanisms, SmartCooper has a lower computational complexity. Assuming there are $N$ CAVs participating in collaboration, the time complexity of the attention mechanism is $O(N^2)$, whereas SmartCooper has a time complexity of $O(N)$.

\subsection{Overall Architecture}

To combine our proposed methods, we show the overall architecture in Fig. \ref{fig:overall_architecture} and the overall algorithm in Algorithm \ref{alg:smartcooper}. Our proposed SmartCooper is divided into three stages.

\textbf{(1) The first stage is network optimization:}
The network optimization process involves collecting parameters from the actual communication environment to compute the optimized compression ratio $\alpha^{*}$ for both the Adaptive Encoder and Adaptive Decoder. In the case of the Adaptive Encoder and Adaptive Decoder, we have directly implemented the Modulated Autoencoder\cite{yang2020variable}. This framework allows for a trade-off between bitrate and distortion by utilizing trade-off parameters. These trade-off parameters correspond one-to-one to compression ratios. Therefore, by adjusting the compression ratio, we can adapt the trade-off parameters.

\textbf{(2) The second stage is transmission \& judgement:}
Each CAV compresses its own camera's image data and transforms it into a bitstream using the Adaptive Encoder. These bitstreams are then transmitted to the Ego CAV. Upon receiving the bitstreams, the Ego CAV reconstructs them into image data using the Adaptive Decoder. Subsequently, a scoring and decision-making process is carried out by the Judger. Images that surpass a specified threshold, along with the Ego CAV's own camera data, are jointly used as inputs and passed to the Fusion Net.

\textbf{(3) The last stage is fusion:}
We adapt the Fusion Net, i.e., CoBEVT \cite{xu2023cobevt}, which is designed for the fusion of vehicular camera data. It conducts merging data from various vehicles and produces a BEV output.
\label{sec:overall architecture}

\begin{algorithm}[t]
    \caption{SmartCooper: Vehicle Collaborative Perception under Adaptive Fusion}
    \label{alg:smartcooper}
    \small
    \begin{algorithmic}[1]
      \REQUIRE Input ego CAV's image sequence: $I_e$. CAVs' image sequence set: $\{I_i\}$. Adaptive encoders: $E_A$. Adaptive decoders: $D_A$. The channel parameters: $N$, $W$, $P_T$, $H$, $\rho_n$. The volume of data transmitted: $V$. The vehicle coordinates: $\{C_0, ..., C_{N-1}\}$.
      \ENSURE Output BEV prediction results and AP@IoU scores.
      \FOR{$i$ from 0 to $N-1$}
        \STATE Calculate the adaptive compression ratio $\alpha_i^{*}$ by Eq. (\ref{eq: compression ratio});
        \STATE The encoder $E_A$ learns to compress raw image consequence $I_i$, then converts them into bistreams;
        \STATE Current vehicle ${CAV}_i$ transmits bistreams to ego AV;
        \STATE The decoder $D_A$ reconstructs the image sequence and obtain new image consequence $\Dot{I_i}$;
      \ENDFOR
      \FOR{$i$ from 0 to $N-1$}
        \STATE Calculate the perception union of the current vehicle ${CAV}_i$ and ego CAV using coordinates $\{C_0, ..., C_{N-1}\}$;
        \STATE Judger scores new image sequence $\Dot{I_i}$ based on the union of sensing coverage.
      \ENDFOR
      \STATE Judger removes the image sequences whose scores are lower than the threshold, then obtain the qualified image sequence set $\{\Ddot{I_j}\}$.
      \STATE Use fusion net to combine the qualified image sequences $\Ddot{I_j}$ with ego CAV's image sequence $I_e$ to conduct BEV prediction results and AP@IoU scores.
    \end{algorithmic}
\end{algorithm}

\section{Experiments}
In this section, we conduct our experiments to evaluate our proposed scheme. We first set up the experiments under different transmission powers and bandwidths. We then compare the performance of our SmartCooper and several non-adaptive methods, respectively. Finally, we show the IoU results and visualize the performance of collaborative perception with the predicted BEV.

\subsection{Dataset}
We conduct experiments using the CAV simulation platform OpenCOOD and validate our method on OPV2V dataset\cite{xu2022opv2v}, which is collected from the CARLA simulator\cite{dosoviskiy2017carla}. The dataset contains 6 road types, 73 different scenes, 232,913 marked 3D detection bounding boxes. The total point cloud and RGB image data volume exceeds 240 GB.

\subsection{Baselines}
\textbf{C-AOL (Channel-aware only):} This scheme denotes that we only use our communication optimizing scheme and adaptive encoder without the adaptive scoring scheme.

\textbf{TWD (Transmission with distribution):} This scheme is based on the work of X. Lyu\cite{Lyu2022two}, which uses distributed methods for optimization in communication.

\textbf{TWF (Transmission with fairness):} This scheme is based on study\cite{magsino2022roadside}, which mainly distributes communication resources to each sub-channel fairly based on Jain’s Network Starvation Fairness Index.

\textbf{No Fusion:} This scheme does not use any collaborative perception, in which ego CAV only uses the information it perceives and does not receive data from other CAVs.

\subsection{Settings}
Our experimental setup complies with 3GPP standards\cite{anwar2019threegpp}. The basic settings are shown as follows. A vehicle has sensing coverage of up to 70 meters and speeds within the range of 0-50 km/h. The default value of the number of cooperative vehicles is 3, and the upper bound of the number of sub-channels $N$ is 3. Most scenarios in the OPV2V dataset\cite{xu2022opv2v} involve collaboration among three or four CAVs. Therefore, the filtering threshold for the judger mechanism is set as follows: In a scenario with three CAVs, the threshold will eliminate data from CAVs whose normalized $score$ is below 0.4 and whose distance from the Ego CAV exceeds $0.5d_{max}$. In a scenario with four CAVs, the threshold will remove data from CAVs whose normalized $score$ falls below 0.3 and whose distance from the Ego CAV is greater than $0.5d_{max}$. Other parameters are set as follows. The default value of the total bandwidth $W$ is set to 200 MHz, and the default value of the transmission power $P_T$ of each vehicle is set to 8 mW. In addition, the road is a bilateral highway with three lanes for vehicles in each direction and with all vehicles uniformly distributed on the road.
\begin{figure}[t]
    \centering
    \includegraphics[width=0.75\columnwidth]{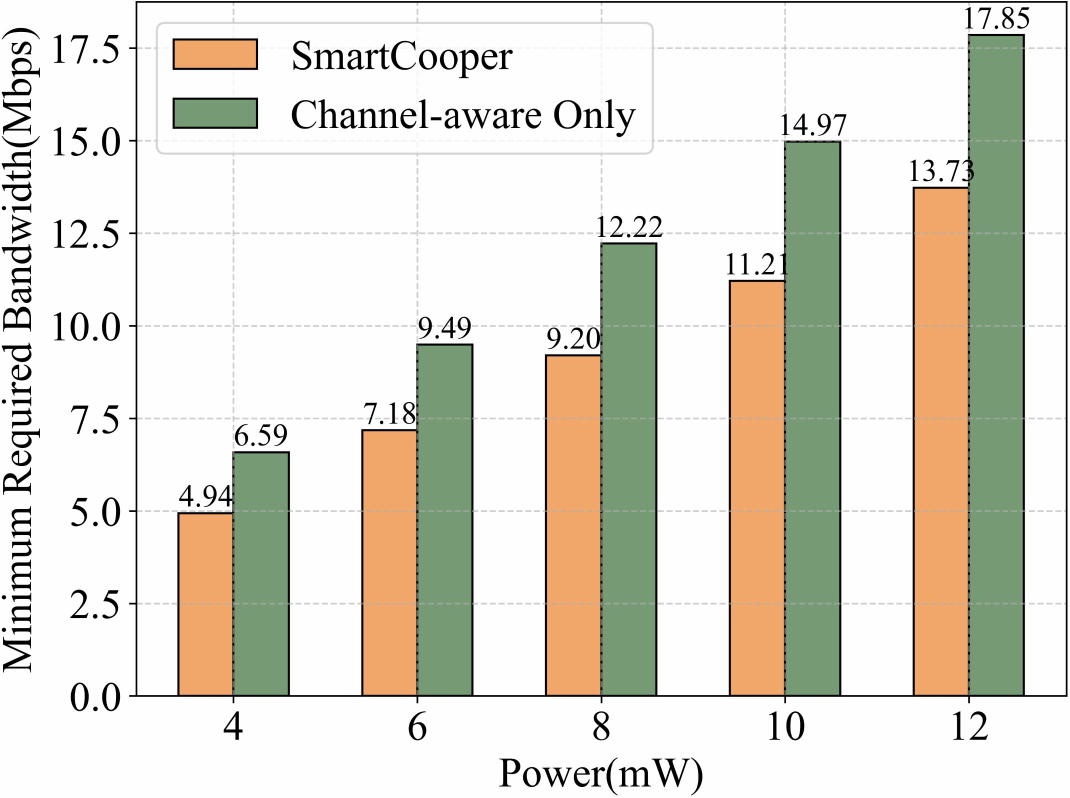}
    \caption{Comparison between SmartCooper and C-AOL for the Minimum Required Bandwidth under different power conditions.}
    \label{fig:minimun_required_bitrate}
    \vspace{-0.3cm}
\end{figure}

\subsection{Experimental Results}

We compare the proposed SmartCooper with four baselines (C-AOL, TWD, TWF and No Fusion). Table \ref{tab:main_result} shows the performance results under different transmission power and bandwidth. The experimental results clearly show that our SmartCooper performs better than all baseline schemes. Specifically, SmartCooper achieves superior AP@IoU scores: in collaborative perception scenes, the AP@IoU scores of SmartCooper outperform TWD by an average of 7.15$\%$ and TWF by 5.65$\%$ under the same power level. Compared with No Fusion scheme, SmartCooper's AP@IoU scores are at least 25.8$\%$ better. When focusing solely on communication optimization, that is, using the C-AOL scheme, the AP@IoU scores outperform TWD by an average of 6.72$\%$, TWF by 5.24$\%$, and No Fusion scheme by 65.3$\%$ under the same power level. Compared with the C-AOL scheme, SmartCooper adds an adaptive judger to filter vehicles at the boundary of the sensing coverage, thereby eliminating the impact of the data with negative gain on data fusion, so the perception performance is better. SmartCooper and C-AOL allocate channel capacity adaptively according to the importance of CAV perception data, while the TWF scheme allocates channel resources fairly, which, in real-world communication environment, is likely to cause some important CAVs to request more communication resources, resulting in poor transmission quality and a deterioration in the collaborative perception. Notice that the data delivered by the TWD scheme is not compressed, which will generate transmission delay under the limited channel capacity. When data of different time scales are fused, the perception results may deteriorate. For No Fusion, SmartCooper and C-AOL integrate information from multiple vehicles, leading to a wider sensing coverage, more comprehensive information, and better perception results than No Fusion.

\begin{table}
\centering
\renewcommand{\arraystretch}{1.4}
\caption{Comparing SmartCooper with four baseline methods across various parameter settings in relation to perception accuracy.}
\label{tab:main_result}
\resizebox{0.49\textwidth}{!}{
\begin{tabular}{|c|ccc|ccc|}
\hline
\multirow{2}{*}{\diagbox[width=10em]{AP@IoU}{Parameters}} & \multicolumn{3}{c|}{\textbf{Power (mW)}} & \multicolumn{3}{c|}{\textbf{Bandwidth (MHz)}}  \\ \cline{2-7}
 & \textbf{4} & \textbf{8} & \textbf{12} & \textbf{100} & \textbf{150} & \textbf{200}  \\ \hline
\textbf{No Fusion} & 0.409 & 0.409 & 0.409 & 0.409 & 0.409 & 0.409   \\ \hline
\textbf{TWD} & 0.600 & 0.633 & 0.669 & 0.669 & 0.669 & 0.669  \\ \hline
\textbf{TWF} & 0.641 & 0.642 & 0.643 & 0.668 & 0.672 & 0.669  \\ \hline
\textbf{C-AOL} & 0.663 & 0.680 & 0.684 & 0.673 & 0.672 & 0.675  \\ \hline
\textbf{SmartCooper} & \textbf{0.667} & \textbf{0.682} & \textbf{0.686} & \textbf{0.686} & \textbf{0.686} & \textbf{0.686} \\ \hline
\end{tabular}
}
\vspace{-0.2cm}
\end{table}

Under practical communication conditions, channel-aware optimization that we propose can significantly improve collaborative perception performance. Furthermore, our proposed SmartCooper not only further enhances collaborative perception quality compared to channel-aware-only optimization but also reduces the minimum required bandwidth for collaboration. In Fig. \ref{fig:minimun_required_bitrate}, SmartCooper reduces the minimum required bandwidth by at least 23.10\% compared to channel-aware optimization. SmartCooper achieves this by intercepting detrimental data to reduce communication overhead while simultaneously improving collaborative perception quality.

\begin{figure}[t]
  \centering
  \subfigure[Groundtruth]{
  \includegraphics[width=3.51cm]{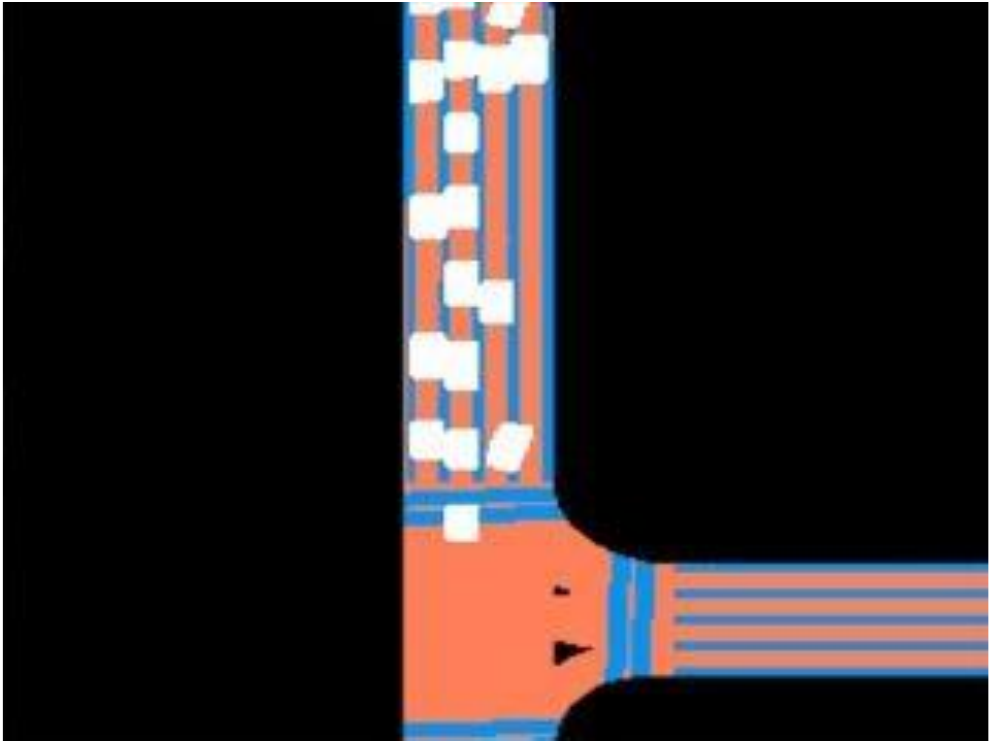}\label{fig:BEV-1}
  }
  \subfigure[SmartCooper]{
  \includegraphics[width=3.51cm]{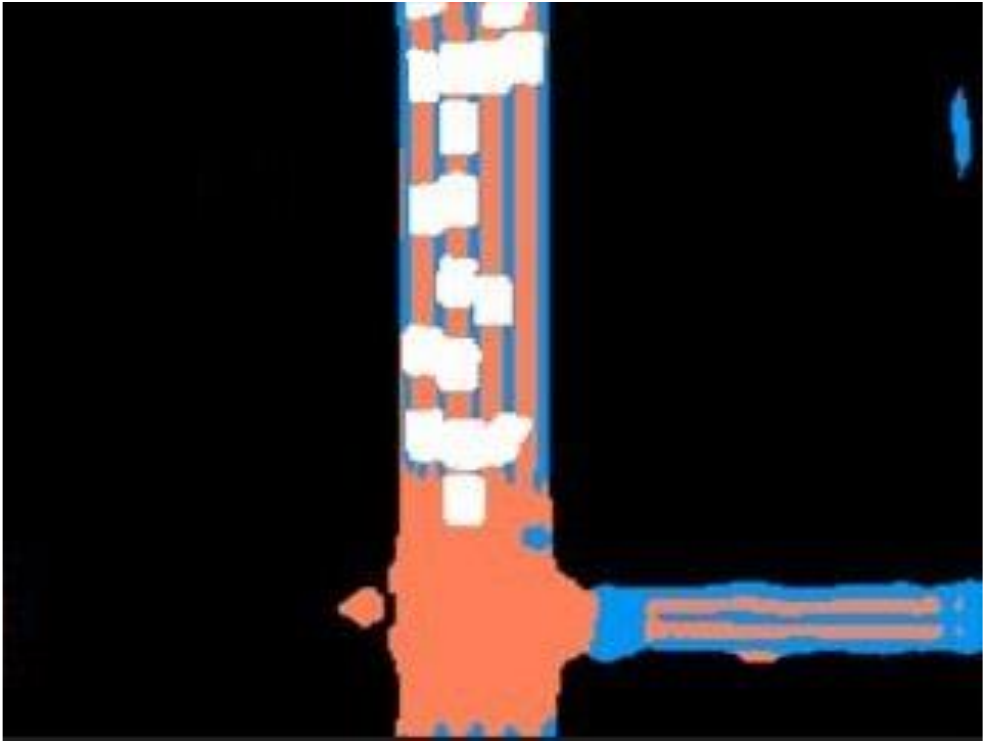}\label{fig:BEV-2}
  }
  \subfigure[C-AOL]{
  \includegraphics[width=3.51cm]{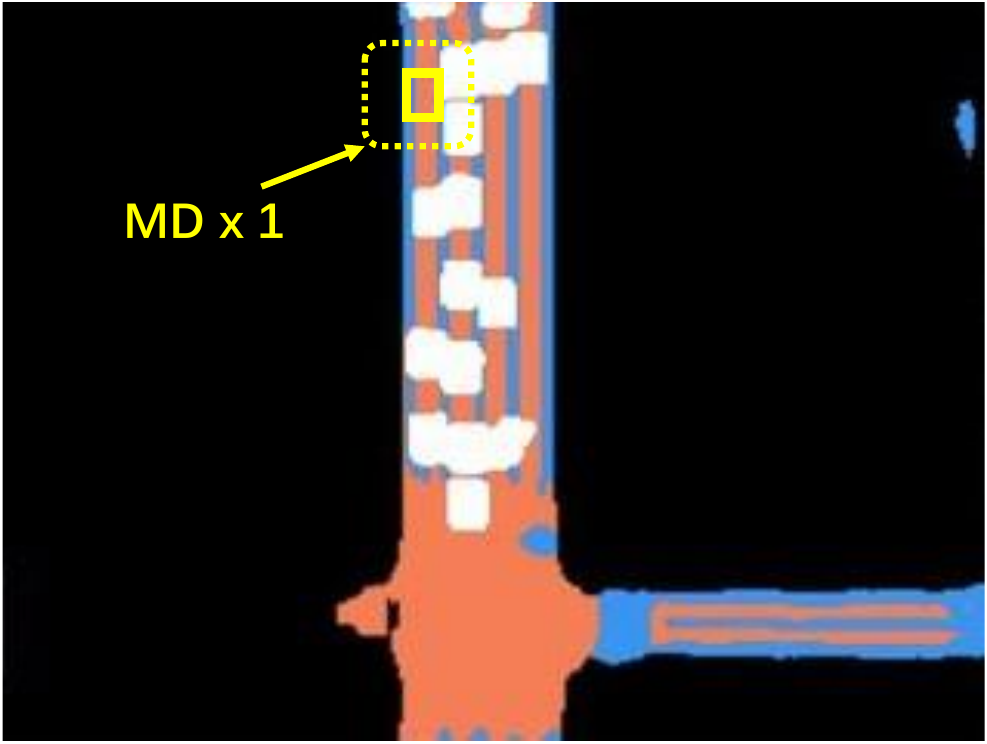}\label{fig:BEV-3}
  }
  \subfigure[TWF]{
  \includegraphics[width=3.51cm]{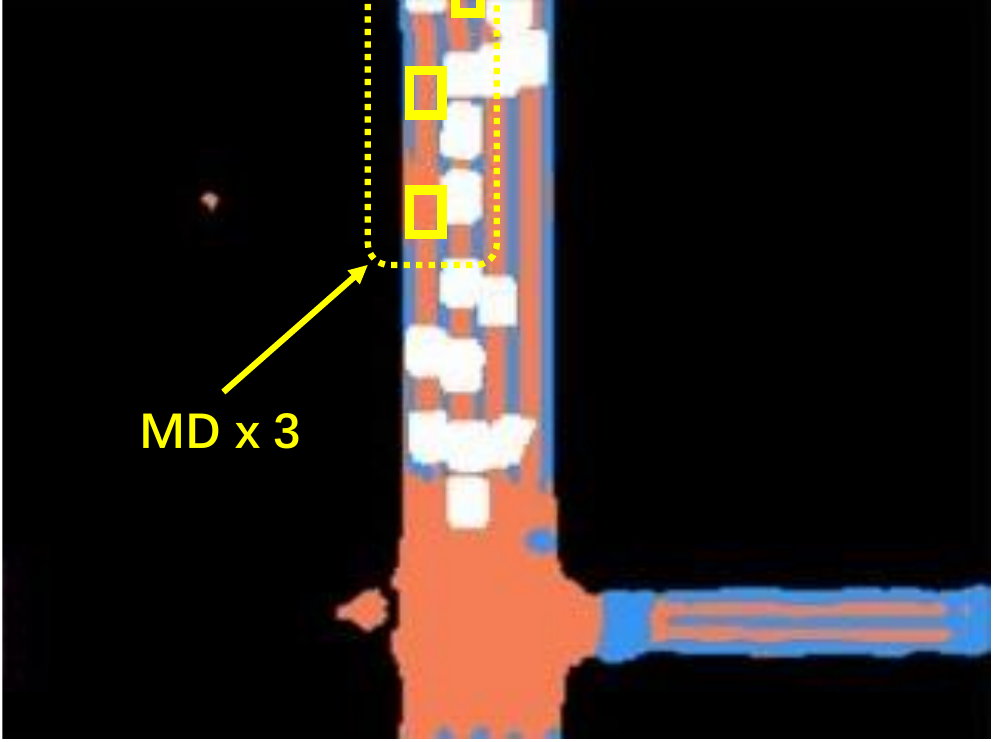}\label{fig:BEV-4}
  }
  \subfigure[TWD]{
  \includegraphics[width=3.51cm]{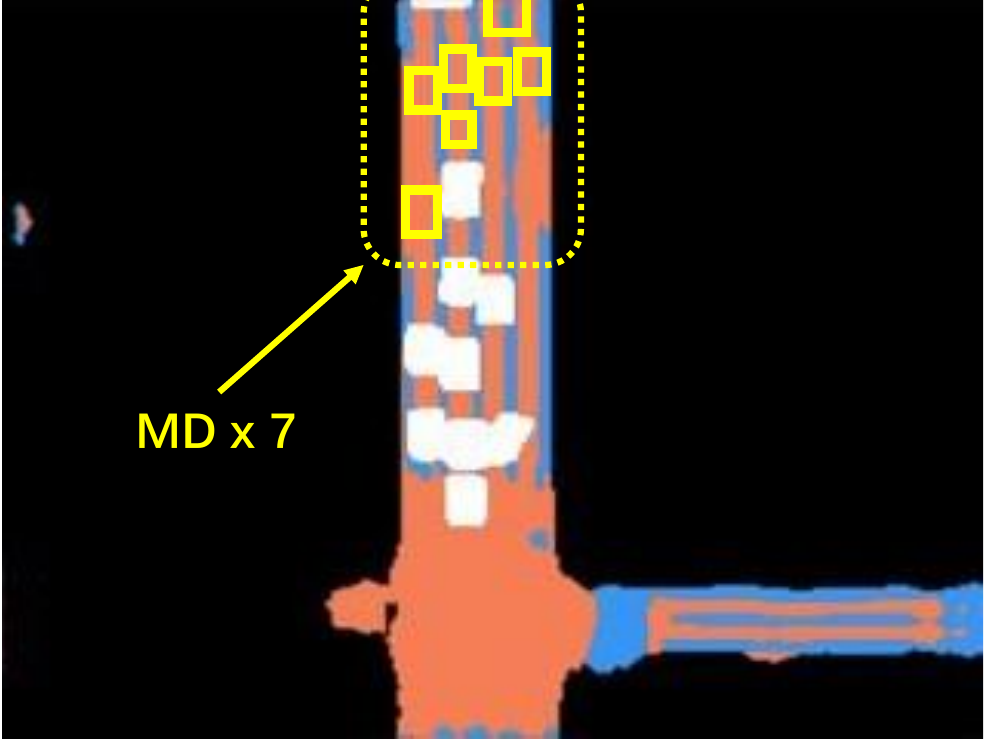}\label{fig:BEV-5}
  }
  \subfigure[No Fusion]{
  \includegraphics[width=3.51cm]{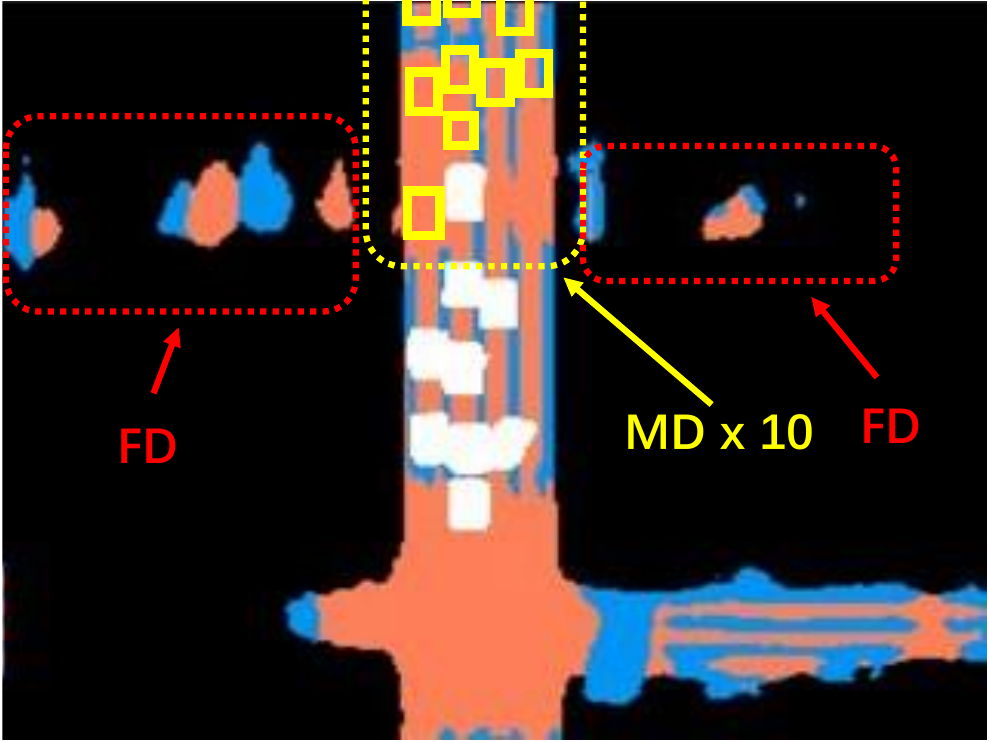}\label{fig:BEV-6}
  }
  \caption{The comparison of SmartCooper with four baseline methods in terms of BEV prediction.}
  \label{fig:bev_results}
  \vspace{-0.4cm}
\end{figure}

Fig. \ref{fig:bev_results} presents the BEV prediction results of (a) Groundtruth, (b) SmartCooper, (c) C-AOL, (d) TWF, (e) TWD and (f) No Fusion at the same time scale and with the same scenario. Groundtruth is the standard result under absolutely ideal condition. For perception schemes, we especially use MD and FD to intuitively denote the errors in prediction results where MD denotes the missed detection, and FD the false detection. Both errors are labeled in each figure. From the visual comparison, our scheme SmartCooper does not have any MD and FD problems and performs the best compared with the baselines. The C-AOL method without Adaptive Scoring mechanism only has one MD problem according to Fig. \ref{fig:BEV-3}. As shown in Figs. \ref{fig:BEV-4} and \ref{fig:BEV-5}, both TWF and TWD have more MD errors. In Fig. \ref{fig:BEV-6}, the No Fusion scheme of single-vehicle detection has not only many MD errors, but also obvious FD errors, which means that it incorrectly predicts roads that do not actually exist. The comparisons of the above results verify the superior performance of our SmartCooper scheme.

\section{Conclusions}
In this paper, we have developed SmartCooper, a scheme aimed at enhancing V2V collaborative perception performance under practical dynamic channel conditions. We first propose C-AOL to ensure collaborative perception quality in actual communication environments by dynamically adjusting compression ratios. However, continuing to increase the number of participating CAVs does not necessarily lead to improved collaborative perception performance under practical channel conditions. Solely relying on channel-aware optimization is insufficient. To solve this problem, SmartCooper employs a distinctive judger mechanism to score and make judgments for each CAV, effectively mitigating the adverse impact of low-quality data on collaborative perception. Experimental results demonstrate that SmartCooper can reduce the communication cost by at least \textbf{23.10\%}. In terms of CAV perception accuracy, SmartCooper outperforms the state-of-the-art schemes by \textbf{7.15\%} under the same transmission power level.


\section*{Acknowledgment}
This work was supported in part by the Hong Kong SAR Government under the Global STEM Professorship and Research Talent Hub and the Hong Kong Jockey Club under JC STEM Lab of Smart City. The work of X. Chen was supported in part by HKU IDS Research Seed Fund under Grant IDS-RSF2023-0012 and HKU-SCF FinTech Academy R\&D Funding.


\newpage
\bibliographystyle{IEEEtran}
\bibliography{ref}

\begin{thebibliography}{10}

\bibitem{anwar2019threegpp}
W~Anwar, N~Franchi, and G~Fettweis.
\newblock Physical layer evaluation of {V2X} communications technologies: {5G}
  {NR-V2X}, {LTE-V2X}, {IEEE} 802.11bd, and {IEEE} 802.11p.
\newblock In {\em IEEE 90th Vehicular Technology Conference (VTC2019-Fall)},
  pages 1--7, Honolulu, HI, USA, Sep. 2019.

\bibitem{caesar2020nuscenes}
H~Caesar, V~Bankiti, A.~H Lang, S~Vora, V.~E Liong, Q~Xu, A~Krishnan, Y~Pan,
  G~Baldan, and O~Beijbom.
\newblock {nuScenes}: A multimodal dataset for autonomous driving.
\newblock In {\em Proceedings of the IEEE/CVF conference on computer vision and
  pattern recognition (CVPR)}, pages 11621--11631, Seattle, USA, Jun. 2020.

\bibitem{chen2019fcooper}
Q~Chen, X~Ma, S~Tang, J~Guo, Q~Yang, and S~Fu.
\newblock F-cooper: Feature based cooperative perception for autonomous vehicle
  edge computing system using {3D} point clouds.
\newblock In {\em Proceedings of the Fourth ACM/IEEE Symposium on Edge
  Computing (SEC)}, pages 88--100, Washington DC, USA, Nov. 2019.

\bibitem{chen2023vaas}
X~Chen, Y~Deng, H~Ding, G~Qu, H~Zhang, P~Li, and Y~Fang.
\newblock {Vehicle as a service (VaaS): Leverage vehicles to build service
  networks and capabilities for smart cities}.
\newblock {\em IEEE Communications Surveys and Tutorials}, 2023 (doi:
  10.1109/COMST.2024.3370169).

\bibitem{cui2022review}
G~Cui, W~Zhang, Y~Xiao, L~Yao, and Z~Fang.
\newblock Cooperative perception technology of autonomous driving in the
  internet of vehicles environment: A review.
\newblock {\em Sensors}, 22(15):5535, Jul. 2022.

\bibitem{dosoviskiy2017carla}
A~Dosovitskiy, G~Ros, F~Codevilla, A~Lopez, and V~Koltun.
\newblock {CARLA}: An open urban driving simulator.
\newblock In {\em Conference on Robot Learning (CoRL)}, pages 1--16, Mountain
  View, CA, USA, Oct. 2017.

\bibitem{fang2022ageofinfo}
Z~Fang, J~Wang, Y~Ren, Z~Han, H.~V Poor, and L~Hanzo.
\newblock Age of information in energy harvesting aided massive multiple access
  networks.
\newblock {\em IEEE Journal on Selected Areas in Communications},
  40(5):1441--1456, May 2022.

\bibitem{geiger2013kitti}
A~Geiger, P~Lenz, C~Stiller, and R~Urtasun.
\newblock Vision meets robotics: The {KITTI} dataset.
\newblock {\em The International Journal of Robotics Research},
  32(11):1231--1237, 2013.

\bibitem{hu2019dnn}
C~Hu, W~Bao, D~Wang, and F~Liu.
\newblock Dynamic adaptive {DNN} surgery for inference acceleration on the
  edge.
\newblock In {\em IEEE Conference on Computer Communications (INFOCOM)}, pages
  1423--1431, Paris, France, Apr. 2019.

\bibitem{hu2023adaptive}
Senkang Hu, Zhengru Fang, Haonan An, Guowen Xu, Yuan Zhou, Xianhao Chen, and
  Yuguang Fang.
\newblock Adaptive communications in collaborative perception with domain
  alignment for autonomous driving.
\newblock {\em arXiv preprint arXiv:2310.00013}, 2023.

\bibitem{hu2023towards}
Senkang Hu, Zhengru Fang, Xianhao Chen, Yuguang Fang, and Sam Kwong.
\newblock Towards full-scene domain generalization in multi-agent collaborative
  bird's eye view segmentation for connected and autonomous driving.
\newblock {\em arXiv preprint arXiv:2311.16754}, 2023.

\bibitem{hu2024collaborative}
Senkang Hu, Zhengru Fang, Yiqin Deng, Xianhao Chen, and Yuguang Fang.
\newblock Collaborative perception for connected and autonomous driving:
  {Challenges}, possible solutions and opportunities.
\newblock {\em arXiv preprint arXiv:2401.01544}, 2024.

\bibitem{hu2022where2comm}
Yue Hu, Shaoheng Fang, Zixing Lei, Yiqi Zhong, and Siheng Chen.
\newblock Where2comm: Communication-efficient collaborative perception via
  spatial confidence maps.
\newblock {\em Advances in neural information processing systems (NeurlPS)},
  35:4874--4886, 2022.

\bibitem{kim2014daev}
S.~W. Kim, B~Qin, Z.~J. Chong, Shen X, W~Liu, M.~H. Ang, E~Frazzoli, and D~Rus.
\newblock Multivehicle cooperative driving using cooperative perception: Design
  and experimental validation.
\newblock {\em IEEE Transactions on Intelligent Transportation Systems (TITS)},
  16(2):663--680, Aug. 2014.

\bibitem{lang2019pointpillars}
A.~H Lang, S~Vora, H~Caesar, L~Zhou, J~Yang, and O~Beijbom.
\newblock Pointpillars: Fast encoders for object detection from point clouds.
\newblock In {\em Proceedings of the IEEE/CVF conference on computer vision and
  pattern recognition (CVPR)}, pages 12697--12705, Long Beach, CA, USA, Jun.
  2019.

\bibitem{zhang2023quality}
Xingyu Li, Yuang Zhang, Yunda Shi, He~Zhu, Jianming Hu, and Lihui Peng.
\newblock Quality assessment of image dataset for autonomous driving.
\newblock In {\em 2023 IEEE International Conference on Imaging Systems and
  Techniques (IST)}, pages 1--6, Copenhagen, Denmark, Oct. 2023.

\bibitem{liy2020lidarreview}
Y~Li, L~Ma, Z~Zhong, F~Liu, M.~A. Chapman, D~Cao, and J~Li.
\newblock Deep learning for lidar point clouds in autonomous driving: A review.
\newblock {\em IEEE Transactions on Neural Networks and Learning Systems},
  32(8):3412--3432, Aug. 2020.

\bibitem{linz2022tracking}
Z~Lin, L~Wang, J~Ding, Y~Xu, and B~Tan.
\newblock Tracking and transmission design in terahertz {V2I} networks.
\newblock {\em IEEE Transactions on Wireless Communications}, 22(6):3586--3598,
  Jun. 2022.

\bibitem{liu2020who2com}
Yen-Cheng Liu, Junjiao Tian, Chih-Yao Ma, Nathan Glaser, Chia-Wen Kuo, and
  Zsolt Kira.
\newblock Who2com: Collaborative perception via learnable handshake
  communication.
\newblock In {\em IEEE International Conference on Robotics and Automation
  (ICRA)}, pages 6876--6883, Paris, France, May 2020.

\bibitem{Lyu2022two}
X~Lyu, C~Zhang, C~Ren, and Y~Hou.
\newblock Distributed graph-based optimization of multicast data dissemination
  for internet of vehicles.
\newblock {\em IEEE Transactions on Intelligent Transportation Systems (TITS)},
  24(3):3117--3128, Dec. 2022.

\bibitem{magsino2022roadside}
E.~R Magsino and I.~W.-H Ho.
\newblock An enhanced information sharing roadside unit allocation scheme for
  vehicular networks.
\newblock {\em IEEE Transactions on Intelligent Transportation Systems (TITS)},
  23(9):15462--–15475, Jan. 2022.

\bibitem{sun2022shift}
T~Sun, M~Segu, J~Postels, Y~Wang, L~Van~Gool, B~Schiele, F~Tombari, and F~Yu.
\newblock {SHIFT}: A synthetic driving dataset for continuous multi-task domain
  adaptation.
\newblock In {\em Proceedings of the IEEE/CVF conference on computer vision and
  pattern recognition (CVPR)}, pages 21371--21382, New Orleans, LA, USA, Jun.
  2022.

\bibitem{talebpour2016review}
A~Talebpour and H.~S Mahmassani.
\newblock Influence of connected and autonomous vehicles on traffic flow
  stability and throughput.
\newblock {\em Transportation research part C: emerging technologies},
  71:143--163, 2016.

\bibitem{wang2020v2vnet}
Tsun-Hsuan Wang, Sivabalan Manivasagam, Ming Liang, Bin Yang, Wenyuan Zeng, and
  Raquel Urtasun.
\newblock {V2VN}et: Vehicle-to-vehicle communication for joint perception and
  prediction.
\newblock In {\em European Conference on Computer Vision}, pages 605--621,
  Glasgow, UK, Aug. 2020.

\bibitem{wei2022threeu}
W~Wei, J~Wang, Z~Fang, J~Chen, Y~Ren, and Y~Dong.
\newblock {3U}: Joint design of {UAV-USV-UUV} networks for cooperative target
  hunting.
\newblock {\em IEEE Transactions on Vehicular Technology}, 72(3):4085--4090,
  Mar. 2022.

\bibitem{xu2018pointfusion}
D~Xu, D~Anguelov, and A~Jain.
\newblock {PointFusion}: Deep sensor fusion for {3D} bounding box estimation.
\newblock In {\em Proceedings of the IEEE conference on computer vision and
  pattern recognition (CVPR)}, pages 244--253, Salt Lake City, UT, USA, Jun.
  2018.

\bibitem{xu2022opv2v}
R~Xu, H~Xiang, X~Xia, X~Han, J~Li, and J~Ma.
\newblock {OPV2V}: An open benchmark dataset and fusion pipeline for perception
  with vehicle-to-vehicle communication.
\newblock In {\em 2022 International Conference on Robotics and Automation
  (ICRA)}, pages 2583--2589, Philadelphia, PA, USA, May 2022.

\bibitem{xu2023cobevt}
Runsheng Xu, Zhengzhong Tu, Hao Xiang, Wei Shao, Bolei Zhou, and Jiaqi Ma.
\newblock {CoBEVT}: Cooperative bird's eye view semantic segmentation with
  sparse transformers.
\newblock In {\em Conference on Robot Learning (CoRL)}, pages 989--1000,
  Atlanta, GA, USA, Mar. 2023.

\bibitem{yang2020variable}
Fei Yang, Luis Herranz, Joost Van De~Weijer, Jos{\'e} A~Iglesias Guiti{\'a}n,
  Antonio~M L{\'o}pez, and Mikhail~G Mozerov.
\newblock Variable rate deep image compression with modulated autoencoder.
\newblock {\em IEEE Signal Processing Letters}, 27:331--335, Jul. 2020.

\end{thebibliography}

\end{document}